# Co-manipulation of soft-materials estimating deformation from depth images

G. Nicola[a,*], E. Villagrossi[a], N. Pedrocchi[a]

[a]*Institute of Intelligent Industrial Technologies and Systems for Advanced Manufacturing, National Research Council of Italy, 20133, Milan, Italy*

**Abstract**

Human-robot manipulation of soft materials, such as fabrics, composites, and sheets of paper/cardboard, is a challenging operation that presents several relevant industrial applications. Estimating the deformation state of the manipulated material is one of the main challenges. Viable methods provide the indirect measure by calculating the human-robot relative distance. In this paper, we develop a data-driven model to estimate the deformation state of the material from a depth image through a Convolutional Neural Network (CNN). First, we define the deformation state of the material as the relative roto-translation from the current robot pose and a human grasping position. The model estimates the current deformation state through a Convolutional Neural Network, specifically, DenseNet-121 pretrained on ImageNet. The delta between the current and the desired deformation state is fed to the robot controller that outputs twist commands. The paper describes the developed approach to acquire, preprocess the dataset and train the model. The model is compared with the current state-of-the-art method based on a camera skeletal tracker. Results show that the approach achieves better performances and avoids the drawbacks of a skeletal tracker. The model was also validated over three different materials showing its generalization ability. Finally, we also studied the model performance according to different architectures and dataset dimensions to minimize the time required for dataset acquisition.

## 1. Introduction

Robots are increasingly assuming the role of assistants to humans in workplaces. However, robots still lack the intelligence to behave as helpful assistants. A paradigmatic application that highlights such limits is collaborative manipulation and transport, which is relevant in multiple fields of application, from industrial logistics, construction, and manufacturing with composite materials to households. For example, in aerospace manufacturing, many production steps require the draping of large patches of carbon fiber. Currently, the transportation of such patches is often manually performed by multiple human workers. On the one hand, it is hard and expensive to automatize the application. On the other hand, fenced industrial robots still lack the dexterity required for such applications. Collaborative manipulation, on the contrary, may allow the achievement of the required level of intelligence and flexibility.

Human-Robot collaborative transportation offers multiple challenges. First, the robot might need to learn the path or the object's final position. The robot infers them based on clear signals from the human operator or via haptic communication through the manipulated object. The manipulation of non-rigid objects, however, limits haptic communication ability. Not all forces and torques applied by the human can be read from the robot end-effector. Second, the robot should share the workload with the human. Still, the human should be able to gain control over the robot, i.e., the robot and human should be able to exchange leader and follower roles continuously [11]. Finally, a manipulation problem comes from the rotation-translation ambiguity, which can arise in human-robot [9] and human-human [14] manipulation.

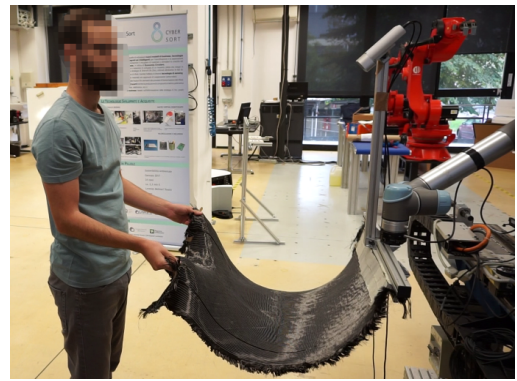

Figure 1: The studied problem of human-robot collaborative transportation of a soft material. The human holds the object from one side and, through an RGB-D camera, estimates the object's deformation.

*Corresponding Author.
Email address: giorgio.nicola@stiima.cnr.it (G. Nicola )

In this paper, we study the problem of human-robot collaborative transportation of soft materials, as shown in Figure 1. Unlike rigid objects, soft materials do not transfer compression forces and display a low stiffness along the other directions. Furthermore, the amount of force/torques they can sustain without damage is minimal. In such a scenario, estimating the deformation state by force measurements is challenging, and standard control architectures based on impedance or admittance control are unfeasible [6, 35].

We propose to use a data-driven approach to develop a Convolutional Neural Network (CNN), which estimates the deformation state of the manipulated soft material from depth images acquired by an RGB-D camera rigidly attached to the robot end-effector. Given a predefined rest configuration state, any estimated variation from such rest state caused by the human can be translated into a movement intention towards the direction that minimizes such difference. The proposed approach is applied to a case of human-robot collaborative transportation of a piece of carbon fiber fabric, proving the method's effectiveness.

The paper is structured as follows: in Section Introduction, related works and paper contributions are presented; in Section Method, the human-robot collaborative transportation problem is formalized, and the proposed solution, including the dataset acquisition strategy, is detailed; In Section Experiments, experimental evaluation of the approach is presented including comparison with state of the art, analysis of different neural network architectures and the dataset dimension; In section Conclusion, conclusions, and future works are highlighted.

### 1.1. Related Work

The manipulation of deformable objects has been deeply investigated [32, 17, 2], and multiple strategies have been developed, sometimes also depending on the geometric and physical characteristics of the objects. Specifically, two main classes of deformable objects are studied: cables [33, 45] and cloth-like [28, 42, 26]. Manipulation of cloth-like objects is typically focused on the cloth folding task. However, few works study the problem of the collaborative human-robot manipulation of such objects [25]; instead, they typically focus on cloth folding [21, 22]. One of the main difficulties in manipulating deformable materials is tracking the object and estimating its current shape, that is, its deformation [38]. Two main approaches have been developed to estimate material deformation in human-robot manipulation of deformable materials: human motion capture and direct deformation estimation. In the first case, the deformation is estimated by tracking the position of the material grasped points by the robot and human by implementing a motion capture system based on either IMU sensors [35] or camera [3]. Tracking the grasping point in the space assumes that the position grasping point on the manipulated object is known as *a priori* or is detectable. Therefore, it is possible to reconstruct the shape of the manipulated material entirely thanks to physics simulation software [19, 3] or directly use the human-robot relative distance as input to the controller [35]. However, reconstructing the manipulated object's shape accurately is a complex task. First of all, it is required to estimate the material parameters, which is a complex and time-consuming activity [5, 36]. Second, it requires high-fidelity simulations that generally cannot be run in real-time. Thus, a trade-off between accuracy and computational load is required. In the second case, the deformation is estimated using a camera to detect visual features on the manipulated material converted into robot commands. Different types of visual features have been developed in the literature. In [18], material folds are detected and combined with force measurements to compute robot speed commands directly. In [15, 16], Histograms of Oriented Wrinkles (HOWs) are implemented as visual features to detect deformations. HOWs are computed by applying Gabor filters and extracting the high-frequency and low-frequency components on the RGB image. Subsequently, in [15], a controller converts the desired speed in the feature to the robot end effector speed. Meanwhile, in [16], a Random Forest controller is trained to compute the desired robot grasping point position from the HOW feature space.

Finally, [6] proposes using motion tracking to detect hand-crafted coded gestures that, combined with torque force measures, are used directly to compute robot end effector speed.

Deep Neural Networks have recently been used as feature extractors via autoencoder networks [46, 37, 40]. The robot's movement to achieve the desired deformation is planned according to the extracted features. However, none of those methods based on DNN feature extraction has been currently applied to human-robot manipulation.

The EU H2020 projects DrapeBot [8] and Merging [27] are pioneering actions coping with these challenges in the industrial scenario. The DrapeBot project focuses on the robotic manipulation of carbon fiber and fiberglass plies during the draping process; in particular, [10] highlights the importance of the human-robot manipulation when the dimension of the ply requires more than one robot. The project Merging looks at the manipulation of flexible and fragile objects exploiting multiple industrial robots, designing new Electro-Adhesive (EA) grasping devices, using information from perception systems fused with the data from a digital twin to estimate the deformations of flexible elements [4, 23].

### 1.2. Contribution

This paper combines the two main literature approaches to estimate material deformation. Specifically, it calculates the relative distance between the robot and human grasping position directly from the material rather than from a motion tracking system. The human-robot relative pose is estimated through a convolutional deep neural network fed with the depth image of the handled material.

Using depth images has multiple advantages. First, motion tracking-based methods assume to know exactly how the human grasps the object because either defined *a priori* or detected. However, such an assumption is only sometimes realistic, especially in unstructured environments such as factories and households. Our method does not track the hands, but it es-timates the robot pose w.r.t. the human hands only by looking at the deformation of the material.

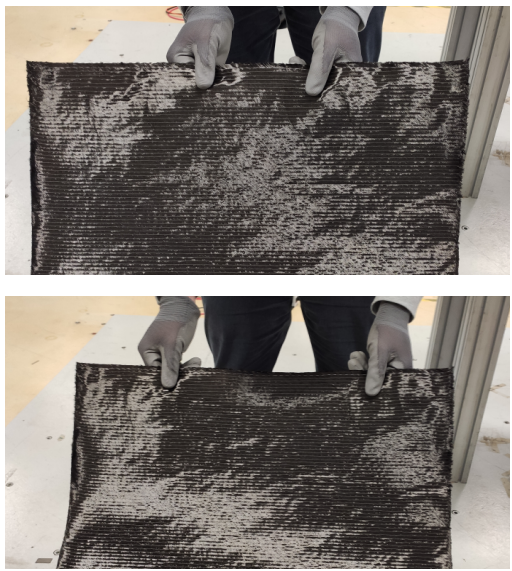

Figure 2: Example of two different human grasping configurations on a carbon fiber ply. The hands do not introduce any deformation on the material between them. The portion of material between them can be considered rigid, and an arbitrary point on the human side of the ply is sufficient to describe its position.

Second, motion tracking-based methods rely on the sensor's performance. On the one hand, IMU-based motion tracking is affected by drift, and it must be worn by the operator either with straps or specific suits. They can be uncomfortable, they can slide, and their position on the body is not repeatable. On the other hand, camera-based motion capture has a narrow working range in distance and field of view. Furthermore, it is sensible for occlusions unless multiple cameras are used.

Third, unlike standard approaches, our approach does not rely on handcrafted features. Indeed, the features might only partially describe the studied problem, and transforming the feature space into robot commands takes a lot of work. Indeed, specific controllers are used, and in some cases, imitation learning on expert user samples is adopted.

Finally, our methodology avoids any use of force feedback. Such sensors are noisy, limited bandwidth, and display low-frequency oscillations given by the object-handled dynamics. Furthermore, the non-collocation of the force/torque sensors introduces translation rotation ambiguity. Specifically, the force and torque sensor measures the resultant of the applied force and torque, but this relation is not bijective. Our approach, instead, uses depth images of the object that are different if the human applies translations or rotations.

The proposed approach is invariant to the soft material properties and independent of the shape. The only assumption is that the depth image of the manipulated material is perceivable. Depending on the application, the depth image can be measured or estimated using different techniques (time of flight with infrared light, stereo camera, single image depth estimation, *etcetera*).

We prove our approach's applicability to collaborative transportation. The robot controller is designed to minimize the distance between the estimated relative pose and a rest relative pose. The paper mainly focuses on the analysis of the performance of deformation estimation. Thus for this applicability test, we decided to prove that, given our formulation, even an extremely simple robot controller can be used differently from the many works in the literature. We leave the in-depth analysis of the different robot controller's performance to future works.

## 2. Method

### 2.1. Problem formulation

The problem of human-robot collaborative transportation of soft materials is composed of two agents handling the soft material simultaneously, as in Figure 1. The human leads the activity, while the robot should follow the human movement. The goal is to minimize deformations from a rest configuration during the manipulation, preserving the material from damage.

Soft materials, like fabric, can be approximated as membranes [41] characterized by the absence of flexural stiffness. Therefore, they cannot sustain compressive loads, and deformations can be caused only by displacements or traction forces. Among the soft materials, we can also distinguish two categories of materials based on their stiffness, whether low, for example, most fabrics, or sufficiently high that elastic deformations cannot be perceived for collaborative transportation. We define the first as non-stiff soft materials and the latter as stiff soft materials.

Let us consider a soft material from now on called ply handled by two agents, a human and a robot. Denote $h_r$ and $h_l$ as the human's hands grasping positions, and $R_{cur}$ as the robot's current grasping point. Given the assumption that the gripper geometry is constant, it is easy to see that a bi-unique proxy for the ply shape is the tuple of the relative roto-translations between the robot grasping position and the human hands holding positions, except for the local deformation around the hands that may change from person to person. This model, however, is unnecessarily accurate. It models the deformation of the ply in the corners, *i.e.*, the roto-translation might change according to the distance human-robot and the grasping positions.

We also assume that the human does not introduce further deformation except those caused by gravity, e.g. pulling apart the hands or folding movements. Such human movements do not have any purpose in a collaborative transportation application. Thus, we can assume that the human operator will not perform them intentionally, and the effect is neglectable if performed. Therefore, the problem formulation can neglect the hands' position and relative distance. Given this assumption, we define an arbitrary point on the ply between the hands invariant to the different grasping positions, see Figure 2. The roto-translation between the robot grasping position and a frame fixed in this point is a proxy for the ply deformation, except for the local deformation around the hands and the geometry of the corners, which are irrelevant to the control objective.

Mathematically, denote $H_{gp}$ as the proxy for the single human grasping point (Figure 3a). Denoting $T_R^H$ and $T_{R,des}^H$ as the actual roto-translation matrix between $R_{cur}$ and $H_{gp}$ and a target desired ply shape. Given this formulation, the problem consists of (i) imposing the target $T_{R,des}^H$ and (ii) estimating $T_R^H$ during the execution movement. The robot should be controlled to minimize the distance from the target's desired pose. Finally, consider the frames as in Figure 3b. We assume that the *x*-axis rotation is always zero since it does not cause macroscopic deformations of the ply but only local deformations.

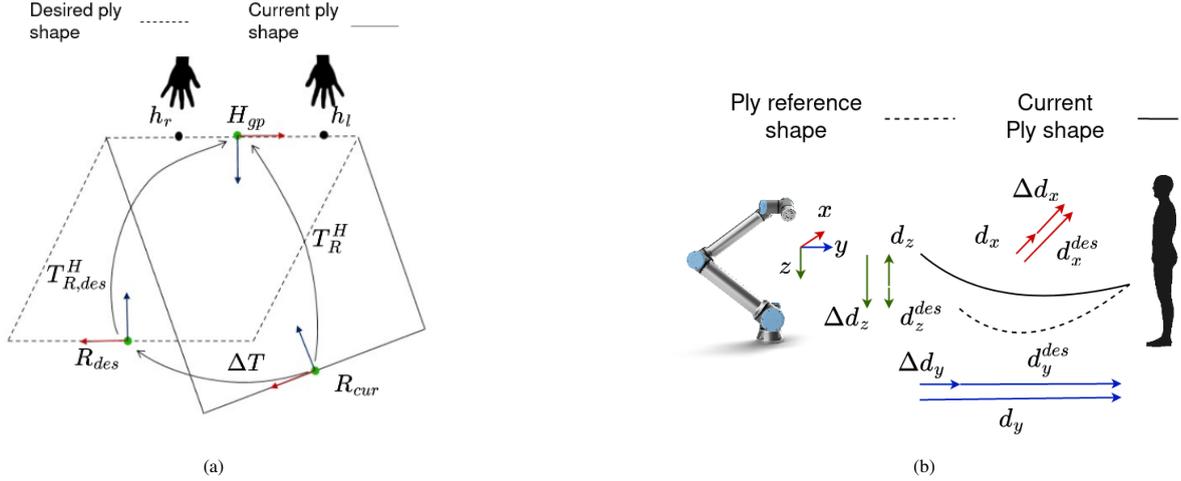

Figure 3: Proposed formalization of the human-robot collaborative transportation problem. a) Shows the top view highlighting the definition of the human grasping point $H_{gp}$. b) Shows a lateral view highlighting the parameters that compose $T_R^H$, $T_{R,des}^H$, and $\Delta T$. For the sake of simplicity, rotations have been neglected.

## 2.2. Proposed solution

We estimate the roto-translation $T_R^H$ through an ensemble of CNNs that takes as input a depth image of the material from a camera rigidly attached to the robot end-effector. Specifically, the ensemble of CNNs outputs the parameters that fully describe $T_R^H$, i.e., three translations and two rotations following the *xyz* Euler conventions. For this purpose, an RGB-D camera is rigidly attached to the robot end-effector that looks at the top of the co-manipulated ply, allowing the easy detection of the material shape and deflections due to the forces/displacement applied by the human on the material (see Figure 1). The 3D camera provides the depth image, and after proper preprocessing, is obtained a depth image composed only of the ply segmented from the background and the human partner. The segmented depth image is fed to an ensemble of CNNs trained to estimate the deformation of the material, in other words, the distance between the robot gripper and the human grasping point $H_{gp}$. Applying Deep Learning techniques allows for defining a black-box model describing the relation between visual deformation and mechanical status. The trained model will likely generalize to any soft material with similar mechanical and visual properties for depth image acquisition. The model will generalize over many stiff soft materials since the elastic deformation is not perceivable for stiff soft materials. Instead, for non-stiff soft materials, the model will likely generalize to materials with the same order of magnitude of stiffness.

## 2.3. Dataset acquisition

The training of the dataset should be done by acquiring a large number of depth images of the deformed material with different human-robot relative distances and human grasping positions. To avoid a bothering effort to a human operator, we substituted the human with a frame that holds the soft material to achieve higher accuracy and repeatability of the dataset, as shown in Figure 4, and the relative distance is got by moving the robot and maintaining the frame fix.

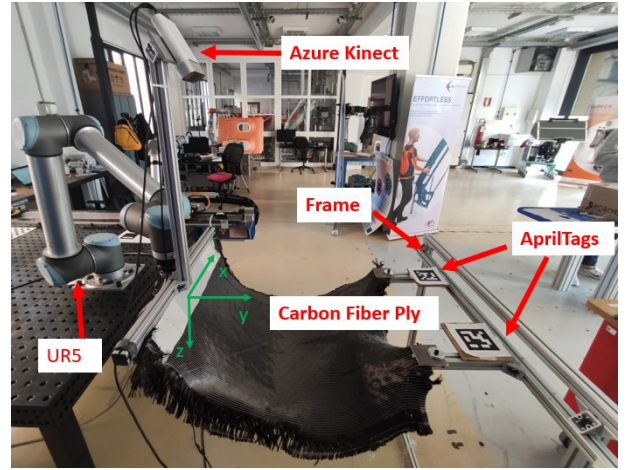

Figure 4: The setup developed to acquire the dataset of a carbon fiber ply. The setup comprises an RGB-D camera, Azure Kinect, a robot Universal Robot UR5, an aluminum frame to mimic the human, and a pair of fiducial markers, Apriltags, to localize the frame.

Precisely, a pair of metallic clips mimic the effect of the hands holding the ply. The frame allows the simulation of different human grasping positions on the material, from now on called human grasping configurations (Figure 2), sliding the beams (Figure 4 see the beams partially covered by the Apriltags) holding the ply on the frame and attaching the clip on the ply accordingly. The frame's position is estimated using a pair of Apriltags [44], and the robot is moved relative to the frame. At each robot position, a set of RGB-D images are taken to account for the camera's noisy output.

The manipulated material is segmented from the background following the scheme shown in Figure 5. First, the depth image is thresholded to a maximum and minimum value, and all pixels whose value is not within this range are set to zero to remove far and close objects quickly. Second, to remove the frame, Apriltags are used, the RGB image is converted into grayscale, and the tags' positions in the 2D picture are detected. Subsequently, tags' positions are used to draw a box, the Apriltag mask, and all pixels outside are set to zero. The Apriltag mask is combined with a static mask identical for each depth image in the dataset. Finally, each image is cropped and resized to the resolution of $224 \times 224$. Each image is autonomously labeled with 3 Cartesian translations and the two rotations describing $T_R^H$ as in Section Problem formulation.

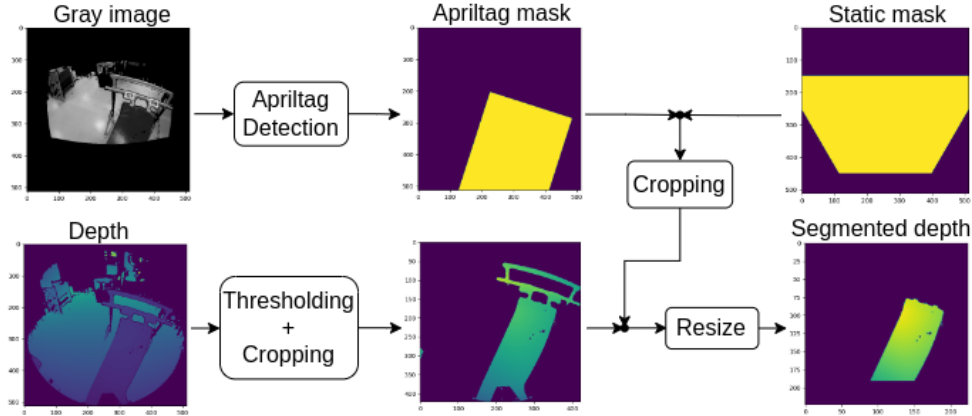

Figure 5: Segmentation pipeline developed for the dataset acquisition. The RGB image converted into a grayscale image is used to compute a mask combined with a static mask. The resulting mask is applied to the depth image after being thresholded.

## 3. Experiments

### 3.1. Experimental setup

We tested the method by manipulating a piece of carbon fiber fabric. Such experimental setup reproduces a relevant aerospace manufacturing application [24, 8]. The carbon fiber piece is $90 \times 60$ cm and thickness 0.45 mm. Given a desired rest configuration of $x_{ref} = 0.0\,m$, $y_{ref} = 0.6\,m$, $z_{ref} = 0.0\,m$, $\theta_{ref} = 0.0°$, $\gamma_{ref} = 0.0°$ the dataset of the deformed material was acquired in the range of $\pm 0.105\,m$ with step $0.03\,m$ on the $x-y-z$ axes and $\pm 20°$ with step $5°$ on both $y$ and $z$ axis rotations for a total of 41472 different poses for each human grasping configuration of the material. We denote the tuple of step values $(0.03\,m, 0.03\,m, 0.03\,m, 5°, 5°)$ used to discretize the acquisition range as the dataset resolution. The dataset resolution defines the maximum permissible estimation error of the roto-translation matrix parameters. The study considered 9 different human grasping configurations, shown in Figure 6, and 373248 labels (combination of robot pose and grasp configuration) for a total of 746496 depth images, *i.e.*, two depth images were taken for each robot pose and grasp configuration. The dataset is available at the following link[1].

[1] https://zenodo.org/record/7871568#.ZEq6Uc5BxEY

The dataset acquisition setup included an RGB-D camera, an Azure Kinect, and a Universal Robot UR5 as a robotic platform. Based on [20], the Azure Kinect, within a range of 2 m, has a spatial systematic error in estimating the depth below 0.002 m that is one order of magnitude below our maximum admissible error of 0.03 m. Thus we consider the Azure Kinect admissible as a sensor for this application.

The pre-processing developed of the acquired dataset is in Section 2.3 and in Figure 5. The depth images were acquired with a wide field of view and binning from the Azure Kinect SDK. No further filtering has been applied either during dataset acquisition or model deployment. The same data augmentation has been applied to all the successive training. Specifically, the data augmentation consisted of random roto-translation, pepper noise, and Poisson noise. The random roto-translations allow the model to be robust against small errors in the camera position. Data augmentation did not include horizontal or vertical flipping, which is commonly used since it can make recognizing between positive and negative rotation impossible. For example, horizontally flipping a depth image with a positive rotation on the z-axis will look the same as with a negative rotation.

### 3.2. Human Robot co-manipulation evaluation

Three CNN models with Densenet-121 [13] compose the system. First, we reserved 1/5 of the dataset for testing. Then each model was trained on a different subset of the remaining dataset using the same principle of K-fold cross-validation with $K = 3$ to generate training and validation datasets for each CNN model. The network hyperparameters have been optimized with the optimization Optuna [1]. Specifically, it implements Bayesian optimization and can prune hyperparameters that are not promising after a few training epochs. We optimized only the learning rate and the batch size since those are the most relevant based on our previous experience [30]. The CNN models are identical except for the input and output layers. The input layer, originally a CNN layer with three channels, was substituted with a CNN layer with one input channel. This design choice is because the proposed method uses only depth images, while Densenet-121 is pretrained on the Ima-geNet dataset [7] composed of RGB images. The weights of the new layer are equal to the sum of the weights along the three original channels. The output layer was replaced with a fully connected layer with 5 outputs, the $T_R^H$ parameters.

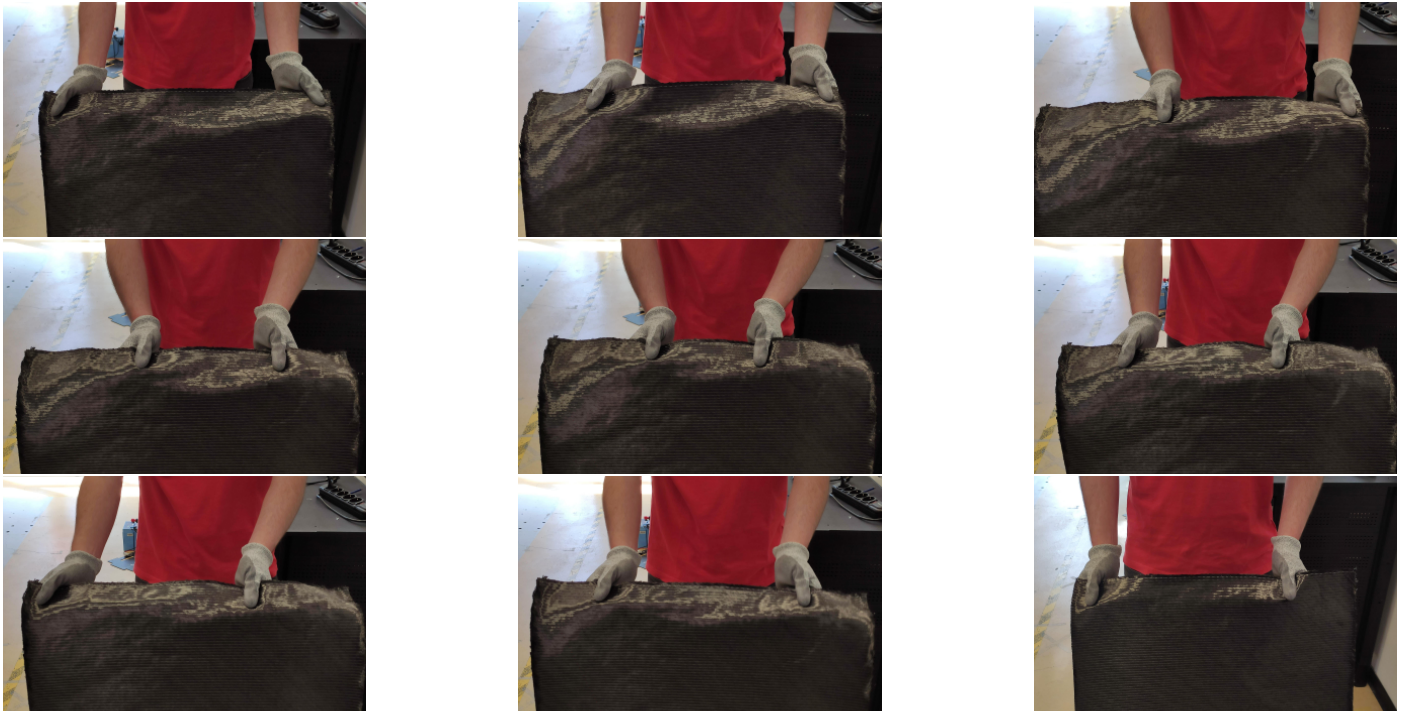

Figure 6: The 9 different grasping configurations studied in the experimental setup.

Figure 7 shows the error distributions of the estimation of the various parameters of $T_R^H$ as boxplots. i.e., the observational error. The error distribution is computed on the entire test dataset and is shown both decomposed in all 5 $T_H^R$ parameters and also as the Cartesian error $E$,

$$E = \sqrt{e_x^2 + e_y^2 + e_z^2}, \quad (1)$$

where $(e_x, e_y, e_z)$ are the translational errors along the 3 axis.

Subsequently, we compared our approach against the commonly used method in literature based on tracking the hands' positions with a skeletal tracker. The hands' key points are used either to compute the hands-robot distance [35] or to perform FEA (Finite Element Analysis) to evaluate the current internal stress state of the object [19, 3]. Both approaches are, at best, as accurate as the skeleton tracker used to track the hands grasping points since they use that information as input. Therefore, we directly compare our approach against the skeleton tracker, and to have a common metric, we compute the human-robot relative distance using the hand key points. The same Azure Kinect is used to acquire depth images of the human and perform skeletal tracking.

Among the many approaches for collaborative manipulation of deformable objects, we compared only with skeletal-tracking based since the other methods do not provide a metric to evaluate the accuracy of deformation estimation. Non-skeletal tracking-based methods directly compute robot commands without first computing the deformation state of the manipulated object. Thus they cannot be used as a comparison to evaluate the observational error of the deformation estimation.

The methods comparison exploited the same setup used for the dataset acquisition, the only difference being that a human was pretending to grasp the deformable material. Given the two hand key points, referring to 2.1, $H_{gp}$ is computed as the midpoint between the two key points. The x-axis is oriented as the segment connecting the two key points and the z-axis is assumed vertical. The depth image segmentation does not use the Apriltags to simulate actual operating conditions for the model. In contrast, it uses the hands' key points positions in the 2D depth image, computed with the skeletal tracker from Azure Kinect, as shown in Figure 9. Specifically, the hands' key points are used to compute a line in the depth image, and all pixels above that line are set to zero. It is essential to note that the skeletal tracker is used for segmenting the depth image only for convenience since developing a method for segmenting the ply is out of the scope of this work. Any method to segment the manipulated material can be implemented based or not on a skeletal tracker. Finally, the robot was positioned in 20 known relative poses to the frame, i.e., set the material with a known deformation state, and we compared both methods on 20 frames for each pose. Results are shown in Figure 8, and the observational error of the two approaches will be discussed, describing the accuracy (mean error) and the precision (standard deviation).

First, we can notice that the accuracy is generally better for the CNN model; the skeletal tracking has a mean error of a few centimeters on all axes and more than 0.1 *rad* on the z-axis rotation. On the contrary, the CNN model has an extremely lower mean error. There are multiple explanations for the lower per-formance of the skeletal tracker. First, the skeletal tracker identifies the hands' key point approximately in the palm's center that might not match exactly with the grasping point with the ply, and it is not realistic to force the human to hold the ply in a specific way. Second, the Azure Kinect skeletal tracker uses as input the depth image, so the skeletal tracker also has to estimate the depth of the key points below the depth image surface. The precision of the CNN model is higher, *i.e.*, lower standard deviation, on the x and z axes. It is similar to the skeletal tracker on the y-axis and y-axis rotation and is lower on the z-axis.

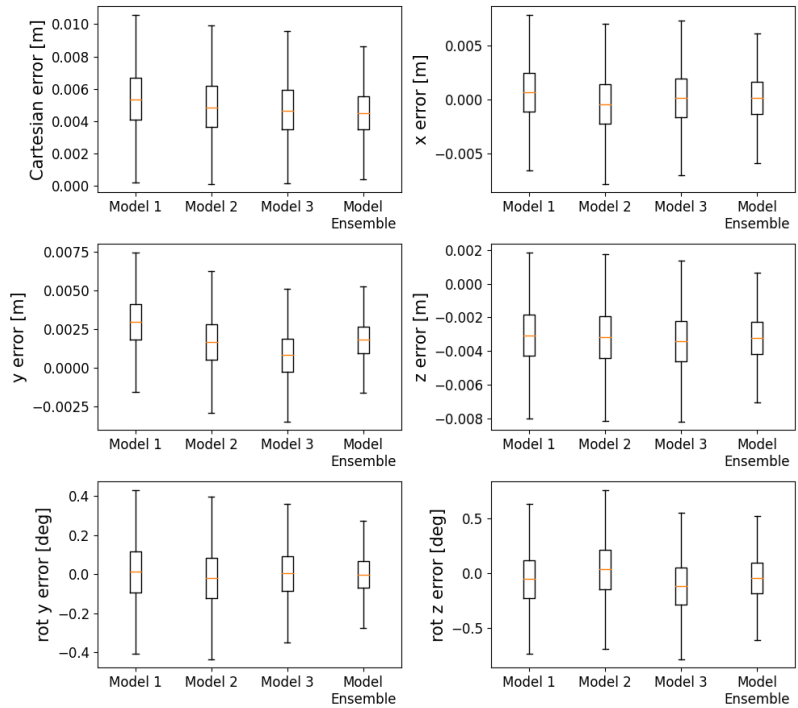

Figure 7: Comparison between the single models and the model ensemble results. **Top left** Cartesian estimation error. **Top right** estimation error on the x-axis. **Center left** estimation error on the y-axis. **Center right** estimation error on the z-axis. **Bottom left** estimation error on the y-axis rotation. **Bottom right** estimation error on the y-axis rotation.

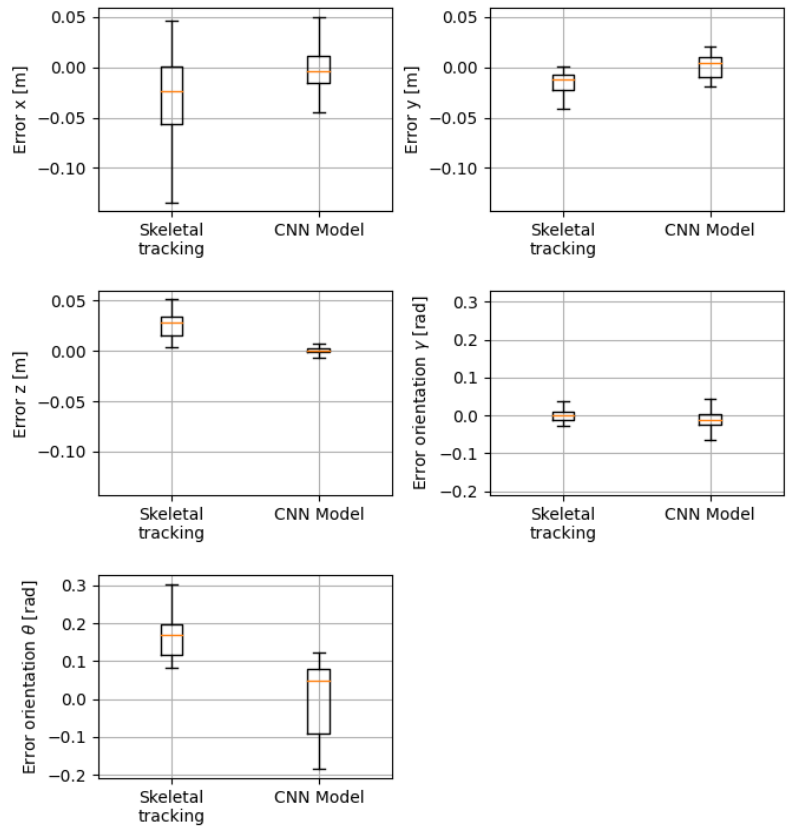

Figure 8: Comparison of the error distributions in estimating the deformation state between the method used in literature based on skeletal tracking and our method based on a CNN model. **Top left** estimation error on the x-axis. **Top right** estimation error on the y-axis. **Center left** estimation error on the z-axis. **Center right** estimation error on y-axis rotation. **Bottom left** estimation error on the z-axis rotation.

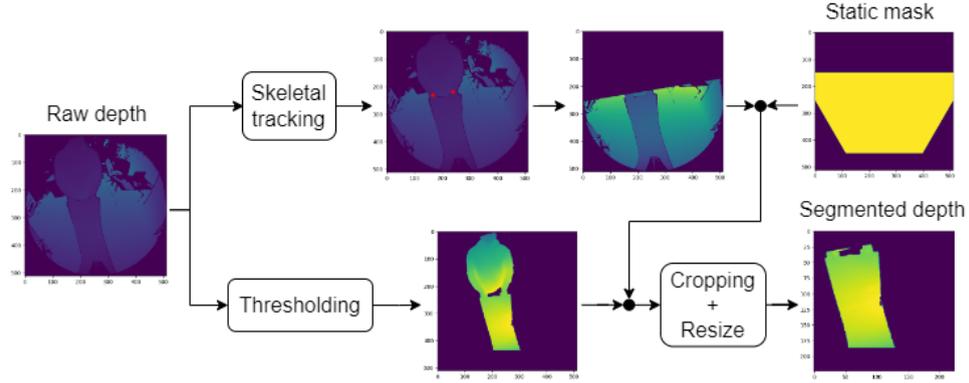

Figure 9: Depth segmentation pipeline developed for collaborative transportation. The pipeline is similar to the dataset acquisition, Figure 5, but the hands' key points from the skeleton tracker are used to trace a line on the depth image instead of computing a mask with Apriltags. Pixels above that line are set to zero.

First, the CNN model estimation errors on the *x* and *z* axes are lower and similar on *y*. The estimation error on the rotations is slightly higher but similar to the skeletal tracking error.

Second, the error of the CNN model compared to the test dataset's error (Figure 7) is significantly higher. The authors believe this difference depends on the various sources of inaccuracies in the experimental setup rather than the model cannot generalize over the different ply segmentation methods. Indeed, the robot and frame grasping positions are hard to reproduce. The same preprocessing pipeline of the dataset acquisition leads to similar results for the model. Even a tiny error in the grasping position, less than 1 *cm*, can significantly change the material shape when highly stretched. Indeed, we noted that the error in the deformation estimation is concentrated in those poses with high values of $\theta$ and $\gamma$ rotations; nevertheless, the rotation was usually underestimated when the ply was highly stretched.

In conclusion, the proposed approach proved to estimate better the *x*, *y*, and *z* positions while slightly worse in the rotation. However, when using the broad field of view, the skeleton tracker method was tested only in optimal conditions, i.e., when the distance camera key points are below 2.5 *m*. Indeed, as described in [39], the skeletal tracking accuracy of the Azure Kinect drastically deteriorates above a threshold distance depending on the used field of view, thus, directly limiting the maximum size of the co-manipulated material.

### 3.3. Network architecture analysis

In this section, we studied the model accuracy using different network architectures. The network architectures studied were the following: a VGG11 without initialized parameters as in [30], VGG11 [34] with batch normalization provided by PyTorch and Densenet-121 [13] provided by PyTorch [31].

The learning rate is equal once to $1 \times 10^{-4}$, and batch size is 256 on three different validation sets. The same conditions were applied to each combination of architecture and dataset dimension to evaluate them in the same situation. After the first 5 epochs without improvements, the learning rate was divided by 10; otherwise, the training was stopped. The maximum number of epochs was 45. For each network, 3 independent networks have been performed each time randomly splitting the dataset: 16/25 training, 4/25 validation, and testing 1/5. Figure 10 shows the training results of each architecture on different percentages of the dataset, including loss and distributions of the Cartesian distance error and orientation error over the y and z axis. Loss is computed as weighted mean squared error, and the Cartesian error is computed as Eq. 1.

$$Loss = \frac{1}{N}\left[\alpha\left(e_x^2 + e_y^2 + e_z^2\right) + \beta\left(e_\theta^2 + e_\gamma^2\right)\right]. \quad (2)$$

The parameters $\alpha$ and $\beta$ are weights introduced in the loss function to balance between errors with different units of measure, specifically meters and radians. It is worth noting that VGG11 and Densenet121 achieve far better results on the test dataset than the VGG not pretrained. This result confirms our hypothesis that the feature learned on the ImageNet dataset is still beneficial even though the input image type differs.

Subsequently, each architecture is trained on a different percentage of the dataset to verify whether a reduced-dimension dataset can still generalize on unseen data. In particular, results are compared between the test dataset and the unused dataset (set of labels removed from the dataset before dividing between training and test dataset). Not surprisingly, for every network architecture, the model performances decrease with the dataset size, with a significant drop at 25%. However, the loss values and error distributions are almost identical between the test and unused sets. It is possible to deduce that 1/4 of the acquired dataset with the deformation range is enough to train a model to generalize over unseen data and specifically unseen deformation combinations. Indeed, in Densenet121 on 25% of the data points in both the test and the unused datasets, the error is far below the dataset resolution (0.03 *m* on the *x-y-z* axis and 5° on $\theta$ and $\gamma$ rotations). However, increasing the dataset size still improves the model's accuracy.

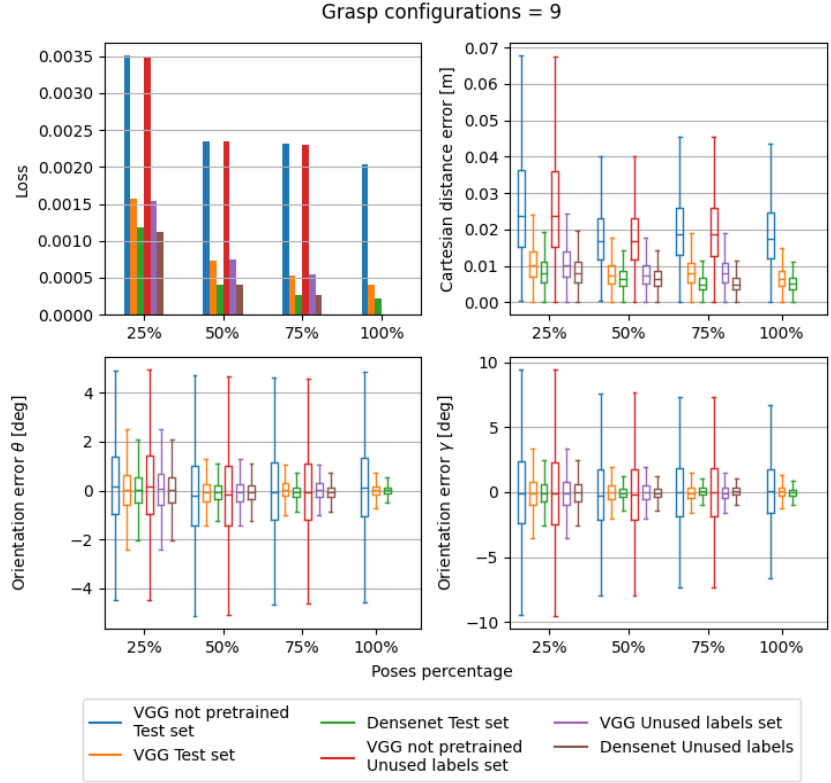

Figure 10: Training results of VGG not pretrained, VGG pretrained, and Densenet121 on the different percentage of the dataset. **Top left** shows the average loss for each architecture on the test and unused datasets. **Top right** shows the Cartesian distance error for each architecture as a boxplot. **Bottom left** shows the orientation error on the y-axis for each architecture as a boxplot. **Bottom right** shows the orientation error on the z-axis for each model as a boxplot.

*3.4. Dataset size analysis*

The dataset size plays an essential role in the training time, which is still a relevant obstacle to implementing Deep Learning methods in the industry.

Three main factors influence the dataset dimension for the presented use case: (i) the number of photos for each pose; (ii) the number of poses for each human grasping configuration; (iii) the number of human grasping configurations.

However, the first point's relevance is minimal since most of the time during the dataset acquisition is spent moving the robot from one pose to another. The photo acquisition takes 0.03*s* while the time point-to-point motion takes 0.2 *s*.

The design of the experiments for study for the dataset size analysis foresees investigations with [100%, 75%, 50%, 25%] randomly sampled over the 41472 poses per grasp configuration and [6, 4] human grasping configurations. Similarly to Section 3.3, the sampled dataset was split as follows: 16/25 training, 4/25 validation, and testing 1/5. Only Densenet-121 was studied since it performed best in all previous training conditions. Each experiment provides the same training conditions for the network architecture analysis.

We compared the results among the test dataset, the "unused labels set", and the "unused grasp set". Specifically, the "unused labels set" includes all the datapoints in the unused set with the grasp configurations on which it is trained. The "unused grasp set" groups all the datapoints with a discarded grasp configuration in the unused set. Table 1 summarizes the dataset size for each training condition.

| N. Grasps | Datataset | Poses percentage | | | |
|---|---|---|---|---|---|
| | | 25% | 50% | 75% | 100% |
| 9 | Train | 59720 | 119439 | 179158 | 238878 |
| | Validation | 14930 | 29860 | 44790 | 59720 |
| | Test | 18663 | 37325 | 55988 | 74650 |
| | Unused labels | 279936 | 186624 | 99312 | 0 |
| 6 | Train | 39813 | 79626 | 149299 | 119439 |
| | Validation | 9953 | 19907 | 29860 | 39813 |
| | Test | 12442 | 24883 | 37325 | 49766 |
| | Unused labels | 186624 | 124416 | 62208 | 0 |
| | Unused grasp | 124416 | 124416 | 124416 | 124416 |
| 4 | Train | 26.542 | 53084 | 79626 | 106168 |
| | Validation | 6635 | 13271 | 19907 | 26542 |
| | Test | 8295 | 16589 | 24883 | 33178 |
| | Unused labels | 124416 | 82944 | 41472 | 0 |
| | Unused grasp | 207360 | 207360 | 207360 | 207360 |

Table 1: Summary of the dataset size during the network architecture and dataset size analysis. Sizes are expressed in labels, combining the robot pose with the grasp configuration. The real number of images is multiplied by the number of photos per robot pose.

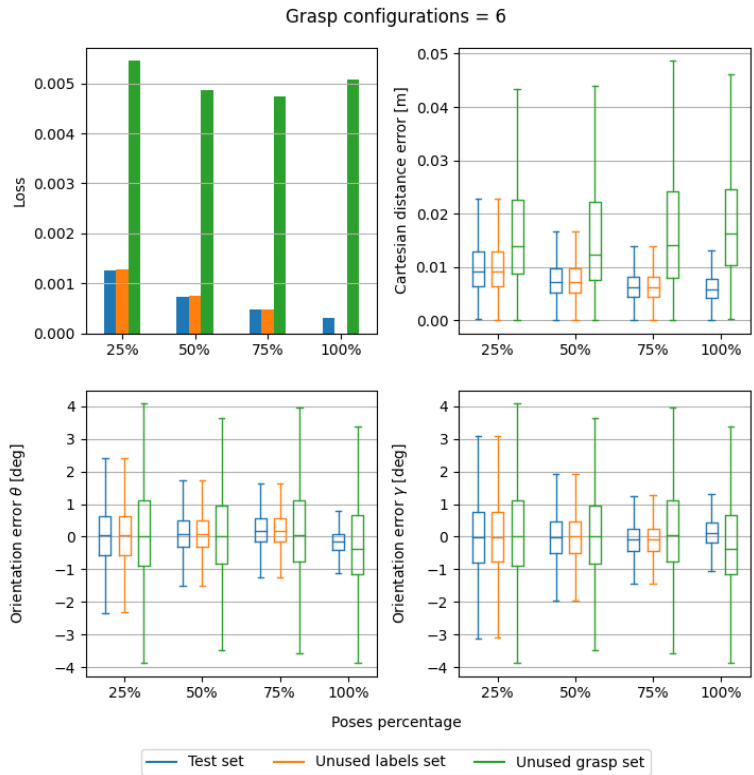

Figure 11: Results training of the DenseNet121 on the different percentages of the dataset and including only 6 grasp configurations over 9, errors are shown as boxplot. **Top left** shows the loss on the test and unused datasets. **Top right** shows the Cartesian distance error. **Bottom left** shows the orientation error on the y-axis. **Bottom right** shows the orientation error on the z-axis.

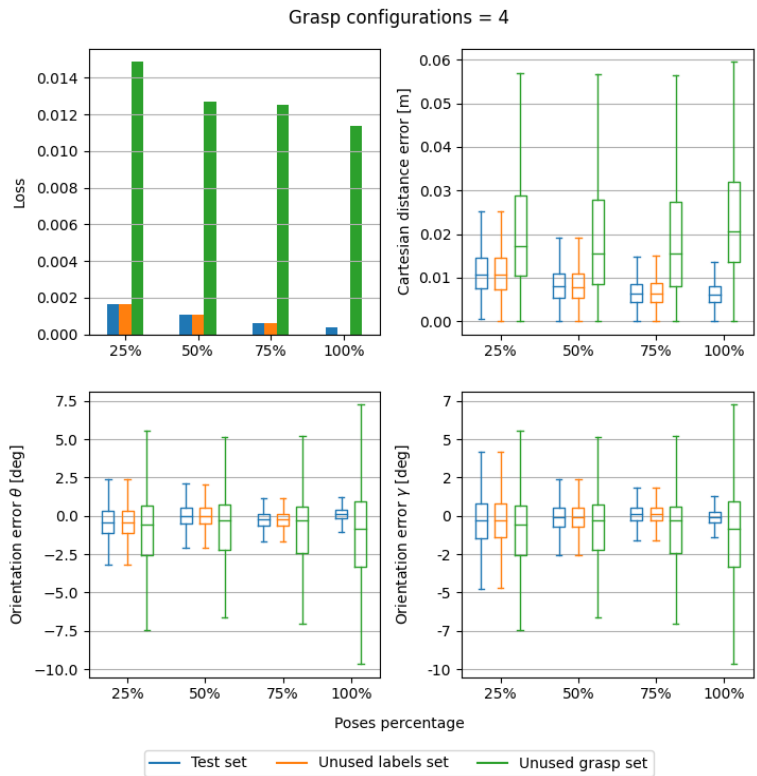

Figure 12: Results training of the DenseNet121 on the different percentage of the dataset and including only 4 grasp configurations over 9. **Top left** shows the loss for each model on the test dataset and the unused dataset. **Top right** shows the Cartesian distance error. **Bottom left** shows the orientation error on the y-axis. **Bottom right** shows the orientation error on the z-axis.

Figure 11 and Figure 12 reports the models results with 6, and 4 grasp configurations. Similarly to the case with all nine grasp configurations, results between the test set and the unused label set are almost identical. On the contrary, there is a significant difference in the results on the unused grasp set. Such results prove that generalizing over unseen grasping configurations is much more challenging than generalizing over unseen poses. Thus, the dataset acquisition should focus on acquiring data with many grasping configurations rather than all the poses. Nevertheless, note that the worst performing model, trained on only 4 grasp configurations and 25% of training poses, estimates the Cartesian position with an error below the dataset resolution in more than 75% of the data points, achieving acceptable results.

### 3.5. Human-robot collaborative transportation

Refer to a scenario of human-robot collaborative transportation of carbon fiber fabric, as shown in Figure 1. The developed model is used to estimate the current deformation state of the carbon fiber fabric, and the delta from a predefined rest deformation state is used to compute robot commands. As described earlier, our approach allows using multiple control strategies. For example, the delta deformation can be directly converted into a twist command with a proportional gain, or it can be converted into a robot tool position reference for a position or impedance control. Otherwise, it can be converted into a virtual force for admittance control. The paper focuses mainly on analyzing the performance of deformation estimation rather than the control strategy we consider a successive problem that should be studied singularly. Thus we decided to prove that effective and natural collaborative co-manipulation can be achieved even with the most straightforward control strategy. In particular, we converted the delta deformation into a robot twist command with a proportional controller, Figure 13, similar to our previous work [30].

A single PC with a CPU, Intel Core i7-8700, and a GPU, NVIDIA GeForce RTX 3080Ti, runs both the model and robot controller. The preprocessing run at approximately 30 Hz, and the robot controller with the model runs at 20 Hz. However, increasing the control rate to 30 Hz is possible using a single CNN model rather than the ensemble. The user could smoothly move the co-manipulated material avoiding excessive material deformations as shown in the video in [29].

### 3.6. Generalization capabilities over different materials

In this section, we studied the capability of a model to generalize over different materials. Two main factors influence the model generalization: 1) material visual properties in the light spectrum used by the sensor for the depth image acquisition; 2) material mechanical properties. The material's visual properties are relevant since using RGB-D cameras very reflective materials beyond certain levels of inclination with respect to the camera tend to be invisible in the depth image. Thus, in very reflective soft materials the depth image has dark spots where the material is very inclined with respect to the camera. Those dark spots essentially are features that the model uses to estimate the deformation status. The material mechanical properties are relevant since the material deformed shape after applying the same deformation status can vary broadly based on the stiffness. Specifically, the deformation due to translations along the $x$ axis is a shear deformation that is significantly affected by the stiffness value.

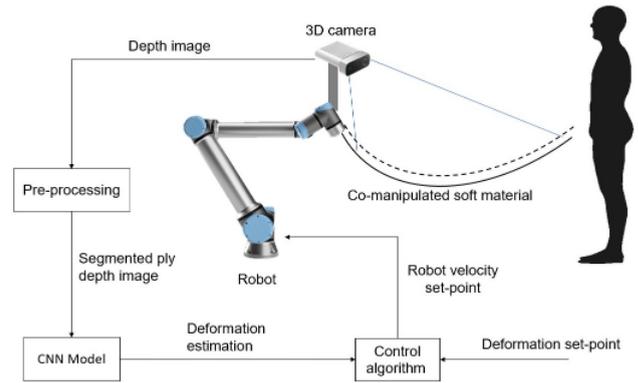

Figure 13: Pipeline developed for human-robot collaborative transportation.

The evaluation consisted in training a model on a material and applying it to different materials, comparing the error distributions. Specifically, the model is trained on prepreg carbon fiber ply (carbon fiber already impregnated with a polymer matrix, from now on called prepreg ply), subsequently tested on a stiffer and heavier prepreg ply (from now on called prepreg stiff ply) and a cotton-made fabric, from now on called cotton ply. Figure 14 shows the developed setup for comparing their error distribution. The prepreg plies can be considered stiff soft materials, while the cotton ply is a non-stiff soft material. The three materials, as shown in Figure 15, are characterized by different levels of reflectivity, with the prepreg stiff ply being the most reflective and the cotton ply being very showing no reflections. Nevertheless, the acquired depth image is very similar among them since depth image based on infrared light is generally less affected by reflections than RGB images.

Following the dataset acquisition pipeline described in Section 2.3 we acquired a dataset including deformations on the $x-y-z$ axes and rotations on the $z$ axis. The ply is 130×38 [$cm$] and has as desired rest configuration of $x_{ref} = 0.0\ m$, $y_{ref} = 1.05\ m$, $z_{ref} = 0.0\ m$, $\theta_{ref} = 0.0°$. The dataset was acquired with a deformation range of ±0.12 [$m$], ±0.255 [$m$], ±0.12 [$m$], ±30° and a step value of 0.03 [$m$], 0.03 [$m$], 0.03 [$m$], 6° and sampling 25% of the robot poses. The model evaluation has been performed by sampling robot poses from a uniform continuous distribution with as support the range of deformation status used during the dataset acquisition.

As shown in Figure 16, the error distribution of the model on the prepreg ply and on prepreg stiff ply is pretty similar and well below the step value used for acquiring the dataset, i.e. the desired accuracy. Instead, the model tested on the cotton ply performs poorly on the *x* axis translations and slightly worse on the *z* axis rotations since the less stiff cotton ply has a very different deformed shape from the prepreg ply. Therefore, we can assert that a model is able to generalize to different materials as long as they have a similar order of magnitude of mechanical properties.

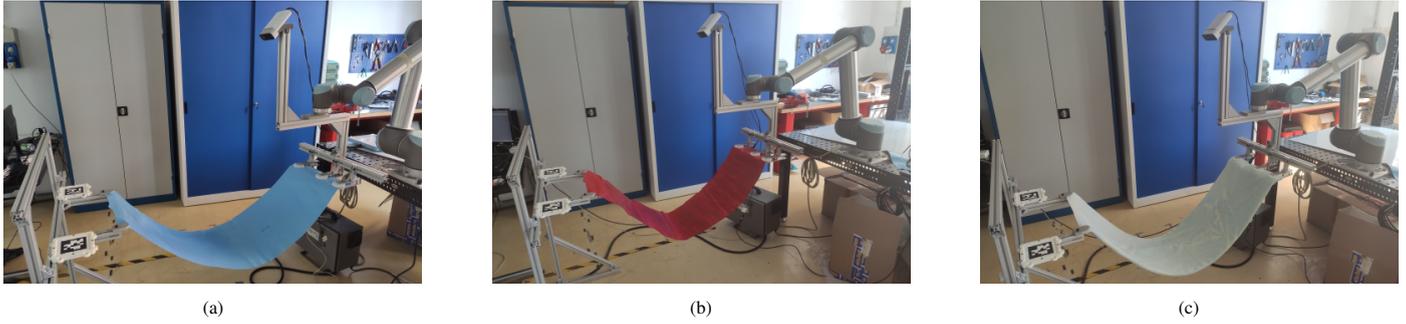

Figure 14: The setup developed to evaluate the capability of the trained model to generalize over different materials. (a) A rectangular ply of prepreg carbon fiber. (b) Stiffer and heavier prepreg carbon fiber ply. (c) A rectangular ply of a cotton-made fabric with the same dimensions

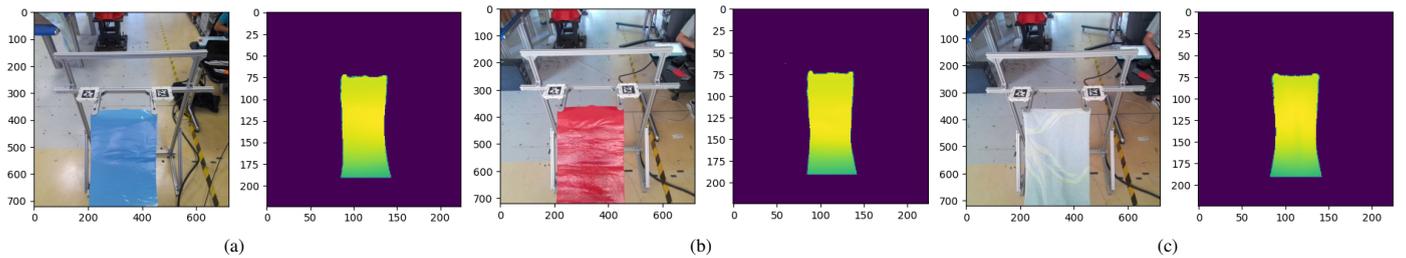

Figure 15: Comparison of the visual properties, left RGB image, right depth image. (a) Prepreg ply. (b) Prepreg stiff ply. (c) Cotton ply.

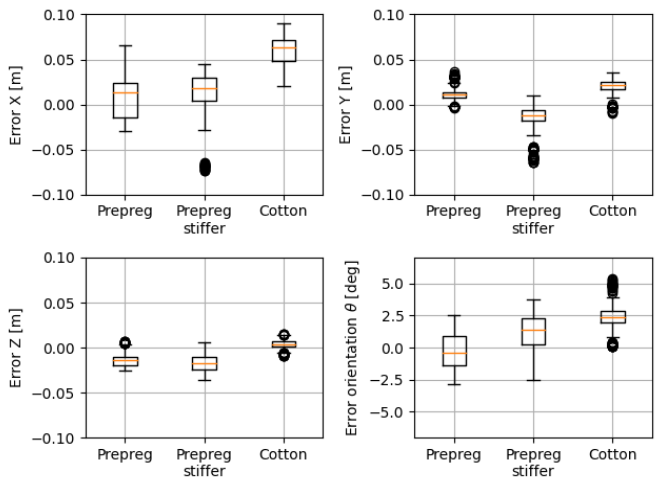

Figure 16: Comparison of the deformation estimation error of the model trained on the prepreg ply on the prepreg and the cotton-made fabric.

In conclusion, the trained model can generalize to other materials only if: 1) they have similar visual properties in the light spectrum used by the camera to acquire the depth image; 2) they have similar mechanical properties.

## 4. Discussion

We compared our approach with estimating the deformation directly from a skeletal tracker. Using a skeletal tracker as input is a common backbone in many works in the literature for collaborative transportation and introduces an additional source of error that cannot be compensated. The skeletal tracker tracks the hands' key points which are used to compute the human-robot relative distance. Demonstrating that the error introduced by the skeleton tracker is higher than the error of our model, we can conclude that the proposed approach is more accurate than the methods in the literature that requires the skeleton tracker as input.

The approach proved to have a lower observational error and is not affected by the two main drawbacks of skeletal trackers. First, the human must always be in the camera's field of view, and second, the accuracy of the skeletal tracker tends to decrease as the human-camera distance increases. Those two drawbacks can have an extremely negative effect when the co-manipulated material size is significant. Meanwhile, our approach can estimate the deformation state of material with any dimension since it only looks at the manipulated material, and all the manipulated material is not required to be within the camera's field of view. Based on the camera's field of view, the ply size, and the level of deformation, part of the ply can extend beyond the camera's field of view. However, our approach can still estimate the deformation status since those are conditions experienced during the dataset acquisition and training. Instead, using a skeletal tracking method, both hands must be visible. Finally, our method is also potentially robust to occlusions since it can be easily trained by introducing artificial occlusions in the data augmentation phase, a common practice in computer vision.

We evaluated multiple network architectures from the literature and studied the model performances according to dataset dimension. Indeed, one of the main limitations to applying Deep Learning models in the industry is the necessity of acquiring large datasets, which is time-consuming. Results showed that the dataset could be acquired with a much lower resolution than the desired maximum estimation error. On the other hand, results also showed that the number of grasping configurations in the dataset is critical since reducing them causes a significant drop in model performances.

The approach was tested in a real-world application of human-robot collaborative transportation of a carbon fiber fabric. The robot was able to follow human movements smoothly, avoiding excessive deformations.

Finally, the model's capability to generalize over different materials was studied. The model trained on prepreg carbon fiber ply was tested on a cotton-made fabric. It was proven that a model could generalize over objects with the same shape and different materials as long as the acquired depth image is similar and the macroscopic mechanical properties are similar. Specifically, soft materials can be divided into two main categories based on whether the stiffness is neglectable. A model can generalize over different materials with similar order of magnitude of mechanical properties.

The main drawback of the proposed approach is that it needs quite a large dataset for every manipulated object that, even if it can be acquired chiefly autonomously, is still time-consuming. Indeed, the model's capability to generalize over different materials is limited and depends on the mechanical and visual properties of the handled material. Models trained on stiff soft materials can generalize well over other stiff soft materials, while non-stiff soft materials are likely to generalize only to materials with similar stiffness. Nevertheless, it should be remembered that other methods in the literature [19, 3] that use physics simulators to estimate the current shape or the internal stress state require a first step of estimating the material mechanical parameters as in [5, 36]. Such activity is complex and time-consuming.

## 5. Conclusions

This work presented a data-driven approach to manipulating soft deformable materials based on estimating the material deformation from depth images from an RGB-D camera rigidly attached to the robot end-effector through a CNN model. First, we formalized the problem of human-robot co-manipulation. We defined the material deformation state based on the roto-translation between the robot and the human grasping points. Second, we developed a Deep Learning model to estimate the roto-translation matrix parameters from a depth image of the deformed material. Results showed that the approach is more accurate than skeletal trackers as input data methods. The experimental evaluation included: the analysis of different network architectures, the study of the required dataset size, and the study of the capability to generalize over different materials. Finally, the approach was tested in a real case of human-robot collaborative transportation.

The main limitation of the approach is the limited capability to generalize depending on the object's properties. Therefore, in many applications, acquiring a new dataset is necessary and even though it is performed autonomously, it is time-consuming. In future works, the authors will investigate how to minimize such limitations, and multiple complementary solutions will be investigated. First, it will be studied the usage of synthetic datasets to train the model. Second, we will train multiple models simultaneously, one for each ply shape and material, sharing a standard backbone for the CNN part of the network. Thus, the models should learn standard and possibly more general features that allow transfer learning by retraining the last fully connected layers. The authors also plan to compare different network architectures taking inspiration from other work in the literature that involves RGB-D data such as pose estimation [43, 12].

This work focuses mostly on the analysis of the deformation estimation; nevertheless, the control strategy is another fundamental step. In the future, the authors plan to investigate different control strategies. Finally, the deformation estimation was applied to the case of collaborative transportation with an anthropomorphic manipulator. The authors plan to apply the proposed approach to a case with an anthropomorphic manipulator mounted on top of a mobile platform to increase the technological fallout of the method.

**Acknowledgement**

This project has been funded by the European Union's Horizon 2020 research and innovation program under grant agreement No 101006732, "DrapeBot – A European Project developing collaborative draping of carbon fiber parts."